\documentclass[twocol]{ametsocV5}

\usepackage{amsmath,amsfonts,amssymb,bm}
\usepackage{mathptmx}
\usepackage{newtxtext}
\usepackage{newtxmath}
\usepackage{comment}
\usepackage[utf8]{inputenc}
\usepackage{multirow}

\title{Hurricane Forecasting: A Novel Multimodal Machine Learning Framework}

\authors{L\'eonard Boussioux*}
\affiliation{\small{Operations Research Center, Massachusetts Institute of Technology, Cambridge, MA, USA \\ \textup{leobix@mit.edu}}}

\extraauthor{Cynthia Zeng\thanks{Equal contribution}}
\extraaffil{\small{Operations Research Center, Massachusetts Institute of Technology, Cambridge, MA, USA \\ \textup{czeng12@mit.edu}}}

\extraauthor{Th\'eo Gu\'enais}
\extraaffil{\small{School of Engineering and Applied Sciences, Harvard University, Cambridge, MA, USA \\ \textup{tguenais@fas.harvard.edu}}}

\extraauthor{Dimitris Bertsimas\correspondingauthor{Dimitris Berstimas, dbertsim@mit.edu}}
\extraaffil{\small{Sloan School of Management and Operations Research Center, Massachusetts Institute of Technology, Cambridge, MA, USA \\ \textup{dbertsim@mit.edu}}}

\abstract{
This paper describes a novel machine learning (ML) framework for tropical cyclone intensity and track forecasting, combining multiple ML techniques and utilizing diverse data sources. Our multimodal framework, called Hurricast, efficiently combines spatial-temporal data with statistical data by extracting features with deep-learning encoder-decoder architectures and predicting with gradient-boosted trees. We evaluate our models in the North Atlantic and Eastern Pacific basins on 2016-2019 for 24-hour lead time track and intensity forecasts and show they achieve comparable mean absolute error and skill to current operational forecast models while computing in seconds. Furthermore, the inclusion of Hurricast into an operational forecast consensus model could improve over the National Hurricane Center's official forecast, thus highlighting the complementary properties with existing approaches. In summary, our work demonstrates that utilizing machine learning techniques to combine different data sources can lead to new opportunities in tropical cyclone forecasting.} 

\begin{document}


\maketitle

%
%
\statement
Machine learning techniques have not been fully explored for improving tropical cyclone movement and intensity predictions. This work shows how advanced machine learning techniques combined with routinely available information can be used to improve 24-hour tropical cyclone forecasts efficiently. The successes demonstrated provide a recipe for improving predictions for longer lead times, further reducing forecast uncertainties and benefiting society.

%




\section{Introduction}

A tropical cyclone (TC) is a low-pressure system originating from tropical or subtropical waters and develops by drawing energy from the sea. It is characterized by a warm core, organized deep convection, and a closed surface wind circulation about a well-defined center. 
Every year, tropical cyclones cause hundreds of deaths and billions of dollars of damage to households and businesses \citep{grinsted}. 
Therefore, producing an accurate prediction for TC track and intensity with sufficient lead time is critical to undertake life-saving measures. 

The forecasting task encompasses the track, intensity, size, structure of TCs, and associated storm surges, rainfall, and tornadoes. Most forecasting models focus on producing track (trajectory)
forecasts and intensity forecasts, i.e., intensity measures such as the maximum sustained wind speed in a particular time interval. Current operational TC forecasts can be classified into dynamical models, statistical models, and statistical-dynamical models \citep{nhc2020doc}. Dynamical models, also known as numerical models, utilize powerful supercomputers to simulate atmospheric fields' evolution using dynamical and thermodynamical equations \citep{hwrf2018doc, ecmwf2019doc}. 
Statistical models approximate historical relationships between storm behavior and storm-specific features and, in general, do not explicitly consider the physical process \citep{cliper5, SHIFOR5}.
Statistical-dynamical models use statistical techniques but further include atmospheric variables provided by dynamical models \citep{shipsdemaria}. Lastly, ensemble models combine the forecasts made by multiple runs of a single model \citep{nhc2020doc}. Moreover, consensus models typically combine individual operational forecasts with a simple or weighted average \citep{consensusintensity, hcca, nhc2020doc, nhc}.

In addition, recent developments in Deep Learning (DL) enable Machine Learning (ML) models to employ multiple data processing techniques to process and combine information from a wide range of sources and create sophisticated architectures to model spatial-temporal relationships. Several studies have demonstrated the use of Recurrent Neural Networks (RNNs) to predict TC trajectory based on historical data \citep{rnn2016, gao, rnn2019}. Convolutional Neural Networks (CNNs) have also been applied to process reanalysis data and satellite data for track forecasting \citep{mudigonda, lian2020cnntrack, sophie} and storm intensification forecasting \citep{springer, nasa}.

This paper introduces a machine learning framework called Hurricast (HUML) for both intensity and track forecasting by combining several data sources using deep learning architectures and gradient-boosted trees. 

Our contributions are three-fold:
\begin{enumerate}
    \item We present novel multimodal\footnote{Multimodality in machine learning refers to the simultaneous use of different data formats, including, for example, tabular data, images, time series, free text, audio.} machine learning techniques for TC intensity and track predictions by combining distinct forecasting methodologies to utilize multiple individual data sources.
    Our Hurricast framework employs XGBoost models to make predictions using statistical features based on historical data and spatial-temporal features extracted with deep learning encoder-decoder architectures from atmospheric reanalysis maps.
    
    \item Evaluating in the North Atlantic and Eastern Pacific basins, we demonstrate that our machine learning models produce comparable results to currently operational models for 24-hour lead time for both intensity and track forecasting tasks. 
    
    
    \item Based on our testing, adding one machine learning model as an input to a consensus model can improve the performance, suggesting the potential for incorporating machine learning approaches for hurricane forecasting. 
\end{enumerate}
 
The paper is structured as follows: Section \ref{sec:data} describes the data used in the scope of this study; Section \ref{sec:methodology} explains the operational principles underlying our machine learning models; Section \ref{results} describes the experiments conducted; Section \ref{conclusions} deals with conclusions from the results and validates the effectiveness of our framework. Finally, Section \ref{sec:discussions} discusses limitations and future work needed for the potential operational deployment of such ML approaches.

\section{Data}\label{sec:data}

In this study, we employ three kinds of data dated since 1980: historical storm data, reanalysis maps, and operational forecast data. 
We use all storms from the seven TC basins since 1980 that reach 34 kt maximum intensity at some time, i.e., are classified at least as a tropical storm, and where more than 60 h of data are available after they reached the speed of 34 kt for the first time. Table \ref{tab:TC1} summarises the TCs distribution in each basin included in our data.

\begin{table*}[!h]
\centering
\caption{Number of TCs meeting our selection criteria from the dataset. We show for each basin and storm category: from Tropical Storm (TS) to Hurricanes of category 1 to 5. We also report the total number of 3-hour interval cases we used from each basin.}

\[\begin{aligned}
\begin{tabular}{|l|cccccc|cc|}
\hline \textbf { Basin } & \multicolumn{6}{|c|}{\textbf{TC Category}} & \textbf{Total TC} & \textbf{Total Cases}\\
 & \text { TS } & 1 & 2 & 3 & 4 & 5 & &\\
\hline \\[-0.8em] \text { Eastern North Pacific (EP) } & 109 & 112 & 57 & 59 & 100 & 14 & 451 & 20,970\\
 \text { North Atlantic (NA) } & 108 & 96 & 46 & 42 & 46 & 17 & 355 & 18,468\\
 \text { North Indian (NI) } & 36 & 13 & 10 & 6 & 8 & 1 & 74 & 2,540\\
 \text { South Atlantic (SA) } & 1 & 1 & 0 & 0 & 0 & 0 & 2 & 16\\
 \text { Southwest Indian (SI) } & 179 & 96 & 73 & 71 & 28 & 0 & 447 & 25,538 \\
 \text { Southern Pacific (SP) } & 117 & 76 & 38 & 45 & 15 & 1 & 292 & 13,319\\
 \text { Western North Pacific (WP) } & 422 & 240 & 158 & 128 & 29 & 1 & 978 & 53,148 \\ \hline \\[-0.8em]
 \text { All Basins } & 972 & 634 & 382 & 351 & 226 & 34 & 2,599 & 133,999 \\
\hline
\end{tabular}
\end{aligned}\]
\label{tab:TC1}
\end{table*}

\subsection{Historical Storm Data Set} \label{statdata}
We obtained historical storm data from the National Oceanic and Atmospheric Administration through the post-season storm analysis dataset IBTrACS \citep{ibtracs}. Among the available features, we have selected time, latitude, longitude, and minimum pressure at the center of the TC, distance-to-land, translation speed of the TC, direction of the TC, TC type (disturbance, tropical, extra-tropical, etc.), basin (North-Atlantic, Eastern Pacific, Western Pacific, etc), and maximum sustained wind speed from the WMO agency (or from the regional agency when not available). 
Our overall feature choice is consistent with previous statistical forecasting approaches \citep{SHIPS1994model, shipsdemaria, sophie}. In this paper, we will refer to this data as \textit{statistical data} (see Table \ref{tab:features}).

The IBTrACS dataset interpolates some features to a 3-hour frequency from the original 6-hour recording frequency. It provides a spline interpolation of the position features (e.g., latitude and longitude) and a linear interpolation of the features not related to position (wind speed, pressure reported by regional agencies). However, the WMO wind speed and pressure were not interpolated by IBTrACS and we interpolated them linearly to match the 3-hour frequency.

We processed statistical data through several steps before inputting it into machine learning models. 
First, we treated the categorical features using the one-hot encoding technique: for a specific categorical feature, we converted each possible category as an additional binary feature, with $1$ indicating the sample belongs to this category and $0$ otherwise. We encoded the basin and the nature of the TC as one-hot features.  Second, we encoded cyclical features using cosine and sine transformations to avoid singularities at endpoints. Features processed using this smoothing technique include date, latitude, longitude, and storm direction\footnote{For example, we encoded the latitude value by $ \cos(\pi \cdot \frac{\text{lat}}{180})$ and $\sin(\pi \cdot \frac{\text{lat}}{180})$ and the date value by $\cos(2\pi \cdot \frac{\text{date}}{365})$ and $\sin(2\pi \cdot \frac{\text{date}}{365})$.}.
 
We also engineer two additional features per time-step to capture first-order dynamical effects: the latitude and longitude displacements in degrees between two consecutive steps.  

Finally, the maximum sustained wind speed feature reported can have different averaging policies depending on the specific reporting agency: 1-minute for US basins and 10-minute for other WMO Regional Specialized Meteorological Centres. We adjust all averaging time periods to 1-minute by dividing the 10-minute values by 0.93 as recommended by \cite{guidelines}.  

\begin{table*}
\caption{
List of features included in our statistical data.} \label{tab:features}
\centering 
\resizebox{\textwidth}{!}{%
\begin{tabular}{p{0.12\textwidth}p{0.1\textwidth}p{0.05\textwidth} p{0.08\textwidth} p{0.18\textwidth} p{0.4\textwidth} }
\hline
               Feature &                        Range &      Unit &        Type &             Processing &                                                                                                                                                                        Description \\
\hline
              Latitude &               [-90.000, 90.000] & deg north &   numerical &            spline interpolation by IBTrACS, standardize &                                                                                                                                           Latitude of the center of the hurricane. \\
             Longitude &             [-180.000, 180.000] &  deg east &   numerical &            spline interpolation by IBTrACS, standardize &                                                                                                                                          Longitude of the center of the hurricane. \\
              WMO Wind &                [10, 165] &     knots &   numerical &           linear interpolation, conversion to 1-min, standardize  &                                                               Maximum sustained wind speed from the WMO agency for the current location. \\
          WMO Pressure &              [880, 1022] &        mb &   numerical &            linear interpolation, standardize &                                                                                                                        Wind pressure from the WMO agency for the current location. \\
      Distance to Land &                [0, 4821] &        km &   numerical &            standardize &                                                                                                                                        Distance to land from the current position. The IBTrACS
land mask includes islands larger than 1400 km$^2.$\\
           Storm Speed &                  [0, 69] &     knots &   numerical &            standardize &                                                                                        Translation speed of the system as calculated from the positions in latitude and longitude. \\
       Storm Direction &                 [0, 360] &       deg &   numerical & cosine \& sine encoding &                                                                          Translation direction of the system, as calculated from the positions, pointing in degrees east of north. \\
Storm Displacement Latitude &                [-2.68, 3.13] &       deg &   numerical &            standardize &                                                                                             Engineered feature, indicating latitude change since the last time step (3 hours ago). \\
Storm Displacement Longitude &                [-3.83, 4.28] &       deg &   numerical &            standardize &                                                                                            Engineered feature, indicating longitude change since the last time step (3 hours ago). \\
                 Basin & [NA, EP, WP, NI, SI, SP, SA] &       N/A & categorical &       one-hot encoding &      Basins include: NA - North Atlantic, EP - Eastern North Pacific, WP - Western North Pacific, NI - North Indian, SI - South Indian, SP - Southern Pacific, SA - South Atlantic \\
            Storm Type &         [DS, TS, ET, SS, MX] &       N/A & categorical &       one-hot encoding & Storm types include: DS - Disturbance, TS - Tropical, ET - Extratropical, SS - Subtropical, NR - Not reported, MX - Mixture (contradicting nature reports from different agencies) \\
\hline
\end{tabular}} 
\end{table*} 

\subsection{Reanalysis Maps}\label{reamaps}

In our work, we used the extensive ERA5 reanalysis data set \citep{era5_good} developed by the  European Centre for Medium-Range Weather Forecasts (ECWMF). ERA5 provides hourly estimates of a large number of atmospheric, land, and oceanic climate variables. The data cover the Earth on a 30km grid and resolve the atmosphere using 137 levels from the surface up to a height of 80km.

\begin{figure}
\centering
\includegraphics[width=7cm]{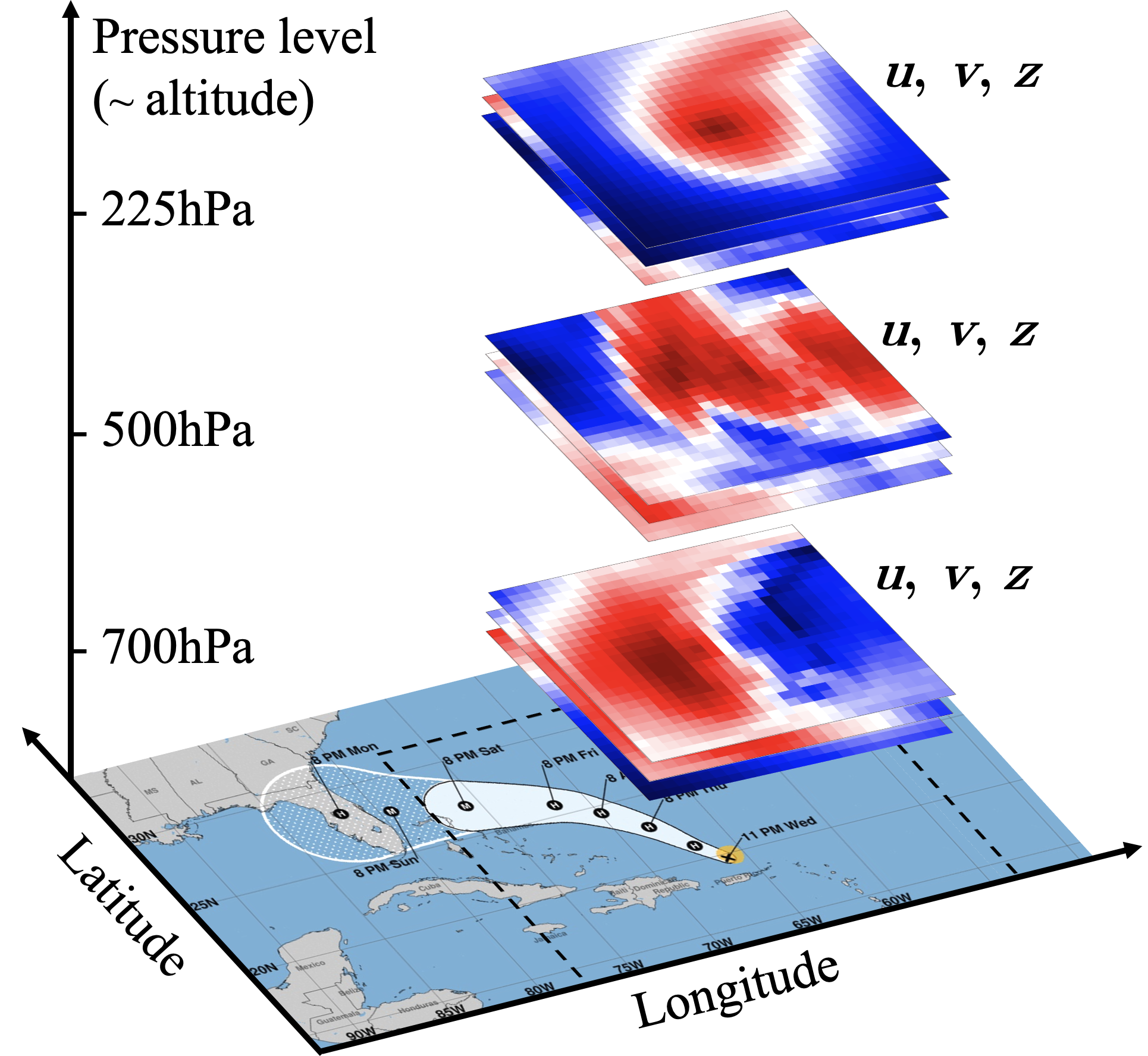}
\caption{Representation of the nine reanalysis maps repeatedly extracted for each time step, corresponding to three different features (geopotential height $z$, zonal component of the wind $u$, meridional component of the wind $v$) at three pressure levels (225, 500, 700 hPa). Each map is of size $25^{\circ}\times25^{\circ}$, centered on the TC center location, and each pixel corresponds to the average field value at the given latitude and longitude degree.} \label{fig:hurricane}
\end{figure}

We extracted $(25\text{°} \times 25 \text{°})$ maps centered at the storm locations across time, given by the IBTrACS dataset described previously, of resolution $1\text{°} \times 1\text{°}$, i.e., each cell corresponds to one degree of latitude and longitude, offering a sufficient frame size to capture the entire storm. We obtained nine reanalysis maps for each TC time step, corresponding to three different features (geopotential height $z$, zonal component of the wind $u$, meridional component of the wind $v$) at three pressure levels (225, 500, 700 hPa), as illustrated in Figure \ref{fig:hurricane}. We chose the three features to incorporate physical information which would influence the TC evolution, and this choice is motivated by previous literature in applying ML techniques to process reanalysis maps \citep{demaria2005, springer, sophie}. 

As a remark, we acknowledge two main limitations from using reanalysis maps for TC forecasting. First, since they are reanalysis products, they are not available in real-time and thus significantly hinder operational use.  Second, they have deficiencies in representing tropical cyclones \citep{schenkel, hodges, bian}; for instance, with large TC sizes particularly being underestimated \citep{bian}.


\subsection{Operational Forecast Models}\label{forecasts}
We obtained operational forecast data from the Automated Tropical Cyclone Forecasting (ATCF) data set, maintained by the National Hurricane Center (NHC) \citep{atcf2, atcf}. The ATCF data contains historical forecasts by operational models used by the NHC for its official forecasting for tropical cyclones and subtropical cyclones in the North Atlantic and Eastern Pacific basins. 
To compare the performance of our models with a benchmark, we selected the strongest operational forecasts with a sufficient number of cases concurrently available: including DSHP, GFSO, HWRF, FSSE, and OFCL for the intensity forecast; CLP5, HWRF, GFSO, AEMN, FSSE, and OFCL for the track forecast (see detailed list in Table \ref{tab:opforecasts}). 
We extracted the forecast data using the Tropycal Python package \citep{tropycal}. 

\begin{table*}

\caption{Summary of all operational forecast models included in our benchmark.}
\centering
\resizebox{\textwidth}{!}{%
$\begin{array}{llll}\hline \text { Model ID } & \text { Model name or type } & \text { Model type } & \text { Forecast } \\ \hline \text { CLP5 } & \text { CLIPER5 Climatology and Persistence } & \text { Statistical (baseline) } & \text { Track }  \\ \text { Decay-SHIPS } & \text { Decay Statistical Hurricane Intensity } & \text { Statistical-dynamical } & \text { Intensity } \\ & \text { Prediction Scheme } & \\ \text { GFSO } & \text { Global Forecast System model } & \text { Multi-layer global dynamical } & \text { Track, Intensity } \\ \text { HWRF } & \text { Hurricane Weather Research and } & \text { Multi-layer regional dynamical } & \text { Track, Intensity } \\ & \text { Forecasting model } & & \\ \text { AEMN } & \text { GFS Ensemble Mean Forecast } & \text { Ensemble } & \text { Track } \\ \text { FSSE } & \text { Florida State Super Ensemble } & \text { Corrected consensus } & \text { Track, Intensity } \\ \text { OFCL } & \text { Official NHC Forecast } & \text { Consensus } & \text { Track, Intensity } \\ \hline\end{array}$}
\label{tab:opforecasts}
\end{table*}

\section{Methodology}
\label{sec:methodology}

Our Hurricast framework makes predictions based on time-series data with different formats: three-dimensional vision-based reanalysis maps and one-dimensional historical storm data consisting of numerical and categorical features. The problem of simultaneously using different types of data is broadly known as multimodal learning in the field of machine learning. 

Overall, we adopt a three-step approach to combine the multiple data sources. We first extract a one-dimensional feature representation (embedding) from each reanalysis maps sequence. 
Second, we concatenate this one-dimensional embedding with the statistical data to form a one-dimensional vector. 
Third, we make our predictions using gradient-boosted tree XGBoost models \citep{xgboost} trained on the selected features.  

\begin{figure}[h] 
    \includegraphics[width=0.95\linewidth]{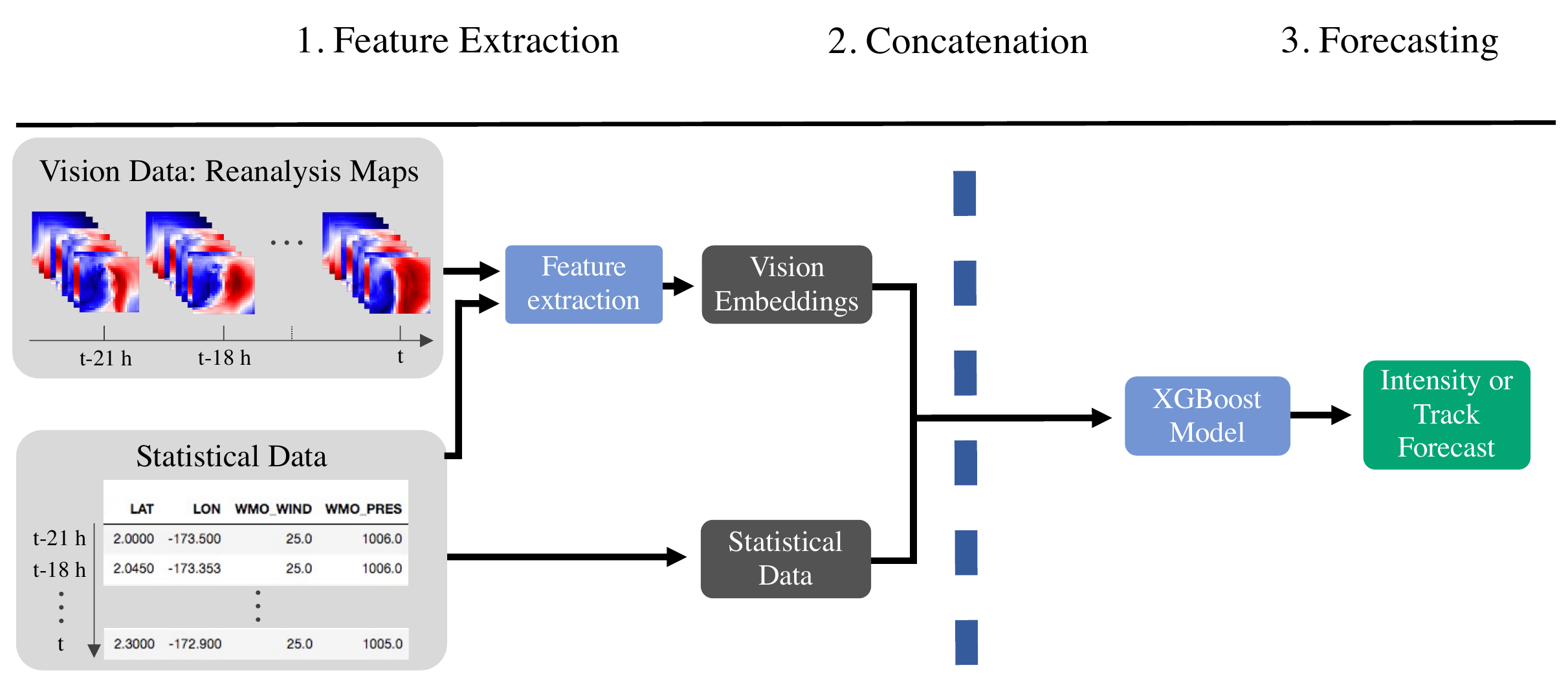}
    \caption{Representation of our multimodal machine learning framework using the two data sources: statistical and reanalysis maps. During Step 1, we extract embeddings from the reanalysis maps. In particular, we use encoder-decoder architectures or tensor decomposition to obtain a one-dimensional representation. 
    During Step 2, we concatenate the statistical data with the features extracted from the reanalysis maps.
    During Step 3, we train one XGBoost model for each of the prediction tasks: intensity in 24 h, latitude displacement in 24 h, and longitude displacement in 24 h. 
    }
    \label{fig:xgb_cnn}
\end{figure}

At a given time step (forecasting case), we perform two 24-hour lead time forecasting tasks: intensity prediction, i.e., predicting the maximum sustained wind speed at a 24-hour lead time; and displacement prediction, i.e., the latitude and longitude storm displacement in degrees between given time and forward 24-hour time. Figure \ref{fig:xgb_cnn} illustrates the three-step pipeline.

To perform the feature extraction in Step 1, we have experimented with two computer vision techniques to obtain the reanalysis maps embeddings: (1) encoder-decoder neural networks and (2) tensor decomposition methods. The former is a supervised learning method; for each input, we use an associated prediction target to train the network. On the other hand, tensor decomposition is an unsupervised method; there is no specific labeled prediction target, and instead, embeddings are drawn directly from the patterns within the data. 

\subsection{Feature Extraction}

\subsubsection{Encoder - Decoder Architectures} \label{sec:cnn-gru}

The encoder-decoder neural network architecture refers to a general type of deep learning architecture consisting of two components: an encoder, which maps the input data into a latent space; and a decoder, which maps the latent space embeddings into predictions. It is well-suited to deal with multimodal data as different types of neural network layers can be adapted to distinct modalities. 

In our work, the encoder component consists of a Convolutional Neural Network (CNN), a successful computer vision technique to process imagery data \citep{cnn1989lecun, imagenet2012, deepnet2016}. 
  
We compare two decoder variations. The first one relies on Gated Recurrent Units (GRU) \citep{gru}, a well-suited recurrent neural network to model temporal dynamic behavior in sequential data. The second one uses Transformers \citep{transformers}, a state-of-the-art architecture for sequential data. While the GRU model the temporal aspect through a recurrence mechanism, the Transformers utilize attention mechanisms and positional encoding \citep{bahdanau2014neural, transformers} to model long-range dependencies.


First, we train the encoder-decoder architectures using standard backpropagation to update the weights parameterizing the models \citep{rumelhart1985learning, deeplearningbook}. We use a mean squared error loss with either an intensity or track objective and add an $L2$ regularization penalty on the network's weights. We then freeze the encoder-decoder's weights when training is completed.

\begin{figure}[h]
    \includegraphics[width=1\linewidth]{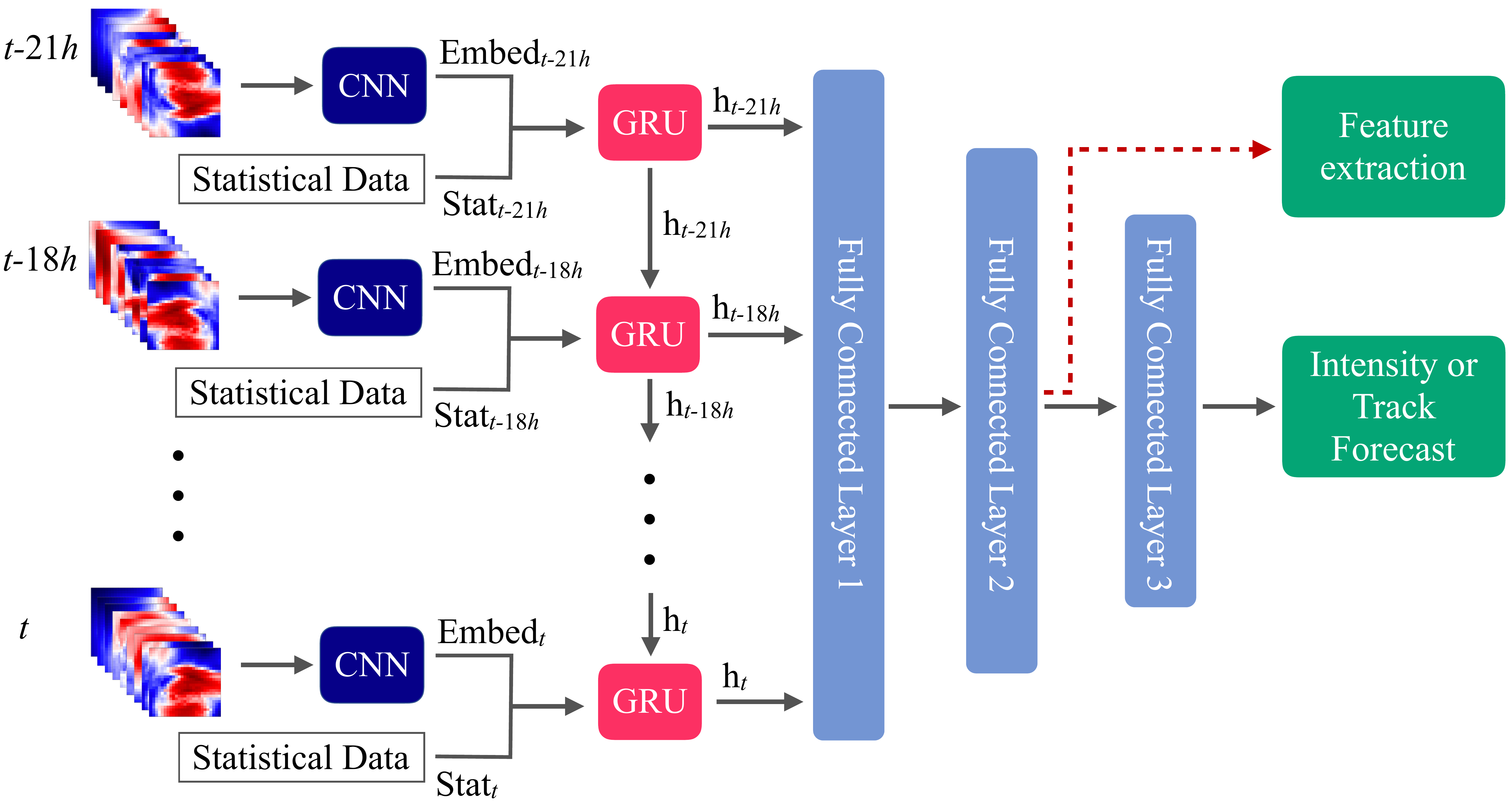}
    \caption{Schematic of our CNN-encoder GRU-decoder network for an 8-time step TC sequence. At each time step, we utilize the CNN to produce a one-dimensional representation of the reanalysis maps. Then, we concatenate these embeddings with the corresponding statistical features to create a sequence of inputs fed sequentially to the GRU. At each time step, the GRU outputs a hidden state passed to the next time step. Finally, we concatenate all the successive hidden states and pass them through three fully connected layers to predict intensity or track with a 24-hour lead time. We finally extract our spatial-temporal embeddings as the output of the second fully connected layer.}\label{fig:cnn_gru}
\end{figure}

\begin{figure}[h]
    \includegraphics[width=1\linewidth]{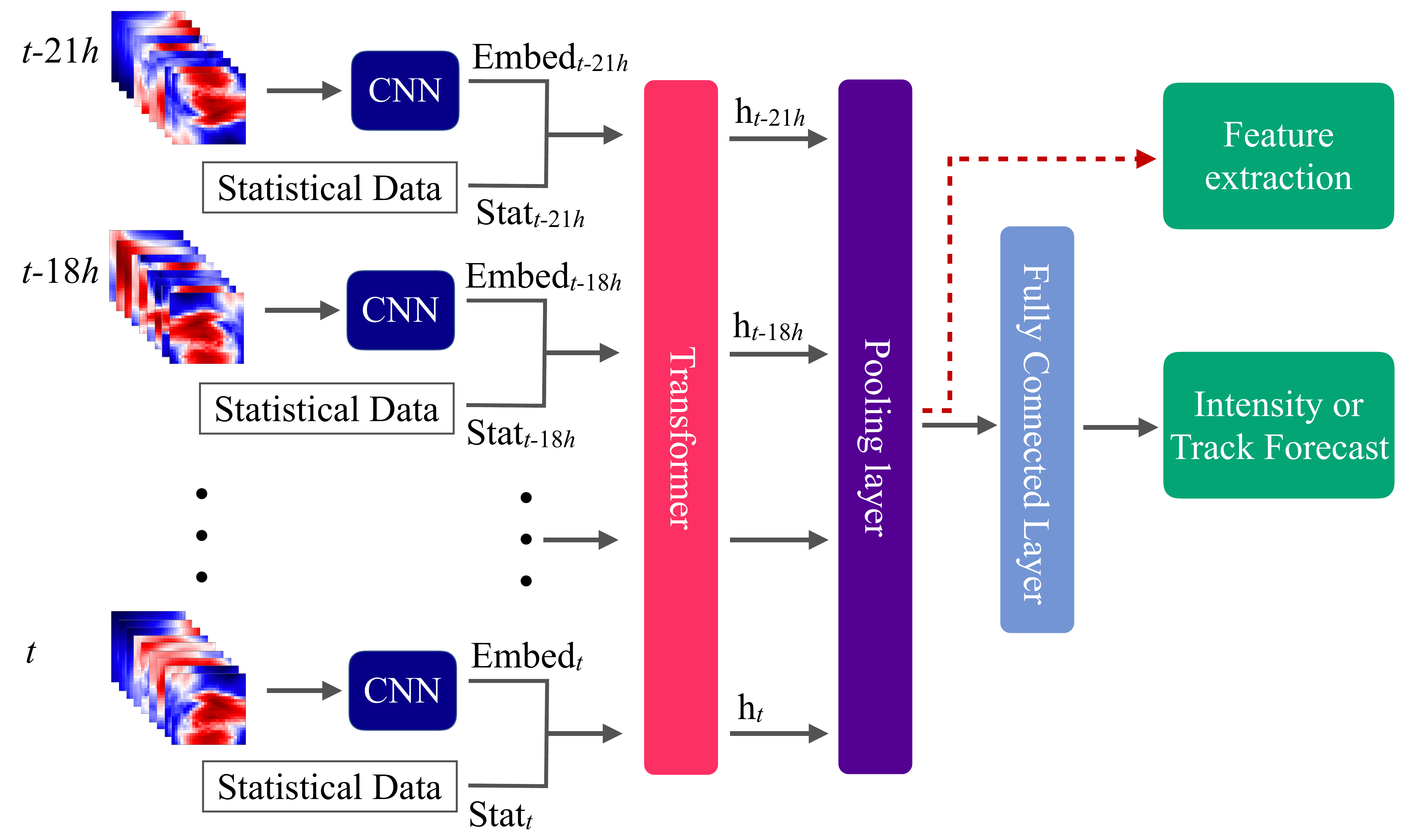}
    \caption{Schematic of our CNN-encoder Transformer-decoder network for an 8-time step TC sequence. At each time step, we utilize the CNN to produce a one-dimensional representation of the reanalysis maps. Then, we concatenate these embeddings with the corresponding statistical features to create a sequence of inputs fed as a whole to the Transformer. The Transformer outputs a new 8-timestep sequence that we average (pool) feature-wise and then feed into one fully connected layer to predict intensity or track with a 24-hour lead time. We finally extract our spatial-temporal embeddings as the output of the pooling layer.}\label{fig:cnn_transfo}
\end{figure}

To perform feature extraction from a given input sequence of reanalysis maps and statistical data, we pass them through the whole frozen encoder-decoder, except the last fully-connected layer (see Figures \ref{fig:cnn_gru} and \ref{fig:cnn_transfo}). The second fully connected layer after the GRU or the pooling layer after the Transformer output a vector of relatively small size, e.g., 128 features, to compress information and provide predictive features. This vector constitutes our one-dimensional reanalysis maps embedding that we extract from the initial 45,000 ($8 \times 9 \times 25 \times 25$) features forming the spatial-temporal input.
The motivation is that since the encoder-decoder acquired intensity or track prediction skills during training, it should capture relevant reanalysis maps information in the embeddings. Using these internal features as input to an external model is a method inspired by transfer learning and distillation, generally efficient in visual imagery \citep{transfer0, transfer1, distillation, transfer2}.

Figures \ref{fig:cnn_gru} and \ref{fig:cnn_transfo} illustrate the encoder-decoder architectures. More details on all components are given in Appendix.

\subsubsection{Tensor Decomposition} \label{sec:tensor}

We also explored tensor decomposition methods as a means of feature extraction. The motivation of using tensor decomposition is to represent high-dimensional data using low dimension features. We use the Tucker decomposition definition throughout this work, also known as the higher-order singular value decomposition. In contrast to the aforementioned neural network-based feature processing techniques, tensor decomposition is an unsupervised extraction technique, meaning features are not learned with respect to specific prediction targets.

At each time step, we treated past reanalysis maps over past time steps as a four-dimensional tensor of size $8 \times 9 \times 25 \times 25 $ (corresponding to 8 past time steps of 9 reanalysis maps of size 25 pixels by 25 pixels). We used the core tensor obtained from the Tucker decomposition as extracted features after flattening it. We decomposed the tensor using the multilinear singular value decomposition (SVD) method, which is computationally efficient \citep{tensor_multisvd}.

The size of the core tensor, i.e., the Tucker rank of the decomposition, is a hyperparameter to be tuned. Based on validation, the Tucker rank is tuned to size $3\times 5\times 3\times 3$. More details on tensor decomposition methodology can be found in the Appendix. 

\subsection{Forecasting Models}

During step 2, we concatenated features from relevant data sources to form a one-dimensional input vector corresponding to each forecasting case.

First, we reshaped the statistical data sequence corresponding to the fixed window size of past observations into a one-dimensional vector. Then, we concatenated it to the one-dimensional reanalysis maps embeddings obtained with one of the feature extraction techniques.

During step 3, we used XGBoost models for the track and intensity forecasts. XGBoost is a gradient-boosted tree-based model widely used in the machine learning community for superior modeling skills and efficient computation time. We compared several other machine learning models during the experimentation phase, including Linear Models, Support Vector Machines, Decision Trees, Random Forests, Feed-forward Neural Networks, and found XGBoost to be generally the most performing. 

\subsection{Summary of Models}\label{models}

This section lists all the forecast models tested and retained and summarizes the methodologies employed in Table \ref{table:methods}. 

\begin{table*}[h]
    \caption{Summary of the various versions of the Hurricast framework for which we report results. Models differ in architecture and data used and are named based on these two characteristics.}
    \centering
    \resizebox{\textwidth}{!}{%
$
\begin{array}{llll}\hline \text{N}^{\circ} & \text { Name } & \text { Data Used } & \text { ML Methods } \\ \hline 
1 & \text { HUML-(stat, xgb) } & \text { Statistical } & \text { XGBoost } \\ 
2 & \text { HUML-(stat/viz, xgb/td) } & \text { Statistical, Vision embeddings} & \text { XGBoost, Feature extraction with tensor decomposition } \\ 
3 & \text { HUML-(stat/viz, xgb/cnn/gru) } & \text { Statistical, Vision embeddings} & \text { XGBoost, Feature extraction with CNN, GRU } \\
4 & \text { HUML-(stat/viz, xgb/cnn/transfo) } & \text { Statistical, Vision embeddings} & \text { XGBoost, Feature extraction with CNN, Transformers } \\ 
5 & \text { HUML-ensemble } & \text { Models 1-4 forecasts} & \text { ElasticNet } \\
6 & \text { HUML/OP-average } & \text { Operational forecasts, } & \text { Simple average } \\
 & & \text{ HUML-(stat/viz, xgb/cnn/transfo)}\\
\hline\end{array}
$}
    \label{table:methods}
\end{table*}


Models 1-4 are variations of the three-step framework described in Figure \ref{fig:xgb_cnn}, with the variation of input data source or processing technique. Model 1, HUML-(stat, xgb), has the simplest form, utilizing only statistical data. Models 2-4 utilize statistical and vision data and are referred to as multimodal models. They differ on the extraction technique used on the reanalysis maps. Model 2, HUML-(stat/viz, xgb/td), uses vision features extracted with tensor decomposition technique. In contrast, Models 3 and 4 utilize vision features extracted with the encoder-decoder, with GRU and Transformer decoders, respectively.
Model 5, HUML-ensemble is a weighted consensus model of Models 1 to 4. The weights given to each model are optimized using ElasticNet. 
Model 6 is a simple average consensus of a few operational forecasts models used by the NHC and our Model 4, HUML-(stat/viz, xgb/cnn/transfo). We use Model 6 to explore whether the Hurricast framework can benefit current operational forecasts by comparing its inclusion as a member model.

\section{Experiments} \label{results}

\subsection{Evaluation Metrics}

To evaluate our intensity forecasts' performance, we computed the mean absolute error (MAE) on the predicted 1-minute maximum sustained wind speed in 24 hours, as provided by the NHC for the North Atlantic and Eastern Pacific basins, defined as:
\[\text{MAE}:=\frac{1}{N} \sum_{i=1}^{N}\left|y_{i}^{\text{true}}-y_{i}^{\text{pred}}\right|,\]

\noindent where $N$ is the number of predictions, $y_{i}^{\text{pred}}$ the predicted forecast intensity with a 24-hour lead time and $y_{i}^{\text{true}}$
the ground-truth 1-min maximum sustained wind speed value given by the WMO agency. 

We computed the mean geographical distance error in kilometers between the actual position and the predicted position in 24 hours to evaluate our track forecasts' performance, using the Haversine formula. The Haversine metric (see Appendix for the exact formula) calculates the great-circle distance between two points --- i.e., the shortest distance between these two points over the Earth's surface. 

We also report the MAE error standard deviation and the forecast skills, using Decay-SHIPS and CLP5 as the baselines for intensity and track, respectively.
 
\subsection{Training, Validation and Testing Protocol} 

We separated the data set into training (80\% of the data), validation (10\% of the data), and testing (10\% of the data). The training set ranges from 1980 to 2011, the validation set from 2012 to 2015, and the test set from 2016 to 2019. Within each set, we treated all samples independently.
 

The test set comprises all the TC cases between 2016 and 2019 from the NA and EP basins where the operational forecast predictions are concurrently available as benchmarks. We compare all models on the same cases.

We use data from all basins during training and validation, but we only report performance on the North Atlantic and Eastern Pacific basins, where we have operational forecast data available.

The precise validation-testing methodology and hyperparameter tuning strategy are detailed in Appendix.

\subsection{Computational Resources}

Our code is available at \url{https://github.com/leobix/hurricast}. We used Python 3.6 \citep{python} and we coded neural networks using Pytorch \citep{pytorch}. We trained all our models using one Tesla V100 GPU and 6 CPU cores. Typically, our encoder-decoders trained within an hour, reaching the best validation performance after 30 epochs.
XGBoost models trained within two minutes.
When making a new prediction at test time, the whole model (feature extraction + XGBoost) runs within a couple of seconds, which shows practical interest for deployment. The bottleneck lies in the acquisition of the reanalysis maps only. We further discuss this point in Section \ref{sec:discussions}.\ref{sec:real-world}.

\section{Results}\label{conclusions}

\paragraph{Standalone machine learning models produce a comparable performance to operational models.\\}

For 24-hour lead time track forecasting, as shown in Table \ref{table:standalone_trk}, the best Hurricast model, HUML-(stat/viz, xgb/cnn/transfo), has a skill with respect to CLP5 of 40\% on the EP basin. In comparison, HWRF has a skill of 45\% and GFSO 46\%. On the NA basin, HUML-(stat/viz, xgb/cnn/transfo) has a skill of 46\%, compared to 63\% for HWRF and 65\% for GFSO. 

For 24-hour lead time intensity forecasting, as shown in Table \ref{table:standalone_int}, the multimodal Hurricast models have a better MAE and lower standard deviation in errors than Decay-SHIPS, HWRF, and GFSO in the EP basin. In particular, our best model, HUML-(stat/viz, xgb/cnn/transfo), outperforms Decay-SHIPS by 12\% and HWRF by 3\% in MAE. 
In the NA basin, HUML-(stat/viz, xgb/cnn/transfo) underperforms Decay-SHIPS by 2\% and HWRF by 7\% but has a lower error standard deviation.

These results highlight that machine learning approaches can emerge as a new methodology to currently existing forecasting methodologies in the field. In addition, we believe there is potential for improvement if given more available data sources.

\begin{table*}[h]
\centering 
\caption{Mean absolute error (MAE), forecast skill with respect to CLP5, and standard deviation of the error (Error sd) of standalone Hurricast models and operational forecasts on the same test set between 2016 and 2019, for 24-hour lead time track forecasting task. Bold values highlight the best performance in each category.}
\begin{tabular}{|c|c|c|c|c|c|c|c|}

\hline & & \multicolumn{3}{|c|}{ Eastern Pacific Basin } & \multicolumn{3}{|c|}{ North Atlantic Basin }\\
Model Type & Model Name & \multicolumn{3}{|c|}{ Comparison on 837 cases }& \multicolumn{3}{|c|}{ Comparison on 899 cases } \\
& & MAE (km) & Skill (\%) & Error sd (km) & MAE (km) & Skill (\%) & Error sd (km) \\

\hline Hurricast & HUML-(stat, xgb) & 81 & 33 & 47 & 144 & 28 & 108 \\

(HUML) & HUML-(stat/viz, xgb/td) & 81 & 33 & 47 & 140 & 30 & 108 \\

Methods & HUML-(stat/viz, xgb/cnn/gru) & $\mathbf{72}$ & $\mathbf{40}$ & $\mathbf{43}$ & 111 & 45 & 79 \\
 & HUML-(stat/viz, xgb/cnn/transfo) & $\mathbf{72}$ & $\mathbf{40}$ & \textbf{43} & $\mathbf{109}$ & $\mathbf{46}$ & $\mathbf{71}$ \\

\hline Standalone & CLP5 & 121 & 0 & 67 & 201 & 0 & 149  \\

 Operational & HWRF & 67 & 45 & 42 & 75 & 63 & \textbf{49} \\

 Forecasts & GFSO & 65 & 46 & 45 & \textbf{71} & \textbf{65} & 54\\

  & AEMN & \textbf{60} & \textbf{50} & \textbf{37} & 73 & 64 & 55\\
 
\hline
\end{tabular}
\label{table:standalone_trk}
\end{table*}

\begin{table*}[!h]
\centering 
\caption{Mean absolute error (MAE), forecast skill with respect to Decay-SHIPS, and standard deviation of the error (Error sd) of standalone Hurricast models and operational forecasts on the same test set between 2016 and 2019, for 24-hour lead time intensity forecasting task. Bold values highlight the best performance in each category.}
\begin{tabular}{|c|c|c|c|c|c|c|c|}
\hline & & \multicolumn{3}{|c|}{ Eastern Pacific Basin } & \multicolumn{3}{|c|}{ North Atlantic Basin }\\
Model Type & Model Name & \multicolumn{3}{|c|}{ Comparison on 877 cases }& \multicolumn{3}{|c|}{ Comparison on 899 cases } \\
& & MAE (kt) & Skill (\%) & Error sd (kt) & MAE (kt) & Skill (\%) & Error sd (kt) \\
\hline Hurricast & HUML-(stat, xgb) & $10.6$ & $9.4$ & $10.5$ & $10.7$ & $-4.9$ & $9.3$ \\
(HUML) & HUML-(stat/viz, xgb/td) & $10.6$ & $9.4$ & $10.4$ & $10.6$ & $-3.9$ & $9.2$ \\
Methods & HUML-(stat/viz, xgb/cnn/gru) & $\mathbf{10.3}$ & $\mathbf{12.0}$ & $10.0$ & $10.8$ & $-5.9$ & $9.2$ \\
 & HUML-(stat/viz, xgb/cnn/transfo) & $\mathbf{10.3}$ & $\mathbf{12.0}$ & \textbf{9.8} & $\mathbf{10.4}$ & $\mathbf{-2.0}$ & $\mathbf{8.8}$ \\
\hline Standalone & GFSO & $15.7$ & $-34.2$ & 14.7 & 14.2 & -39.2 & 14.1  \\
 Operational & Decay-SHIPS & $11.7$ & $0.0$ & $\mathbf{1 0 . 4}$ & $10.2$ & $0.0$ & 9.3 \\
 Forecasts & HWRF & \textbf{10.6} & \textbf{9.4} & 11.0 & \textbf{9.7} & \textbf{4.9} & \textbf{9.0}\\
\hline
\end{tabular}

\label{table:standalone_int}
\end{table*}

\paragraph{ Machine learning models bring additional insights to consensus models.\\}

Consensus models often produce better performance than individual models by averaging out errors and biases.
Hence we conducted testing for two consensus models: HUML-ensemble is the weighted average of all individual Hurricast variations; HUML/OP-consensus is a simple average of HUML-(stat/viz, xgb/cnn/transfo) and the other standalone operational models included in our benchmark. 

As shown in Tables \ref{table:cons_trk} and \ref{table:cons_int}, HUML-ensemble consistently improves upon the best performing Hurricast variation in terms of MAE, showcasing the possibility of building practical ensembles from machine learning models.

Moreover, OP-average consensus is the equal-weighted average of available operational forecasts. We constructed the HUML/OP-average  consensus with the additional inclusion of the HUML-(stat/viz, xgb/cnn/transfo) model. 
Results show that the inclusion of our machine learning model brings value into the consensus for both track and intensity tasks. In addition, HUML/OP-average produces lower MAE and standard deviation under our testing scope than the NHC's official forecast OFCL for 24-hour lead time.

In particular, in our 24-hour lead time testing scope, in terms of intensity MAE, HUML/OP-average outperforms OFCL by 8\% on the EP basin and 2\% on the NA basin. In track MAE, HUML/OP-average outperforms OFCL by 7\% on the EP basin and 14\% on the NA basin.

As a remark, we do not consider the computational time lag of operational model forecasts in our experiments. Computational time varies and can take several hours for dynamical models. Nevertheless, these results highlight the complementary benefits of machine learning models to operational models. 

\begin{table*}[h]
\centering 
\caption{Mean absolute error (MAE), forecast skill with respect to CLP5, and standard deviation of the error (Error sd) of consensus models compared with NHC's official model OFCL on the same test set between 2016 and 2019 for track forecasting task. Bold values highlight the best performance in each category.}
\begin{tabular}{|c|c|c|c|c|c|c|c|}
\hline & & \multicolumn{3}{|c|}{ Eastern Pacific Basin } & \multicolumn{3}{|c|}{ North Atlantic Basin }\\
Model Type & Model Name & \multicolumn{3}{|c|}{ Comparison on 837 cases }& \multicolumn{3}{|c|}{ Comparison on 899 cases } \\
 &  & MAE (km) & Skill (\%) & Error sd (km) & MAE (km) & Skill (\%) & Error sd (km) \\\hline 
Hurricast & HUML-(stat/viz,   xgb/cnn/transfo) & 72 & 40 & 43 & 109 & 46 & \textbf{71} \\
 Methods & HUML-ensemble & \textbf{68} & \textbf{44} & \textbf{41} & \textbf{107} & \textbf{47} & 76 \\\hline 
Operational & FSSE & 56 & 54 & 47 & \textbf{69} & \textbf{66} & \textbf{53} \\
 Forecasts & OFCL & \textbf{54} & \textbf{55} & \textbf{33} & 71 & 65 & 56 \\ \hline 
Consensus& OP-average consensus & 55 & 55 & 37 & 64 & 68 & 48 \\
 Models & HUML/OP-average consensus & \textbf{50} & \textbf{59} & \textbf{32} & \textbf{61} & \textbf{70} & \textbf{42}\\ \hline 
\end{tabular}
\label{table:cons_trk}
\end{table*}

\begin{table*}[h]
\centering 
\caption{Mean absolute error (MAE), forecast skill with respect to Decay-SHIPS, and standard deviation of the error (Error sd) of consensus models compared with NHC's official model OFCL on the same test set between 2016 and 2019 for intensity forecasting task. Bold values highlight the best performance in each category.}
\begin{tabular} {|c|c|c|c|c|c|c|c|}
\hline & & \multicolumn{3}{|c|}{ Eastern Pacific Basin } & \multicolumn{3}{|c|}{ North Atlantic Basin }\\
Model Type & Model Name & \multicolumn{3}{|c|}{ Comparison on 877 cases }& \multicolumn{3}{|c|}{ Comparison on 899 cases } \\
& & MAE (kt) & Skill (\%) & Error sd (kt) & MAE (kt) & Skill (\%) & Error sd (kt) \\\hline 
Hurricast & HUML-(stat/viz,   xgb/cnn/transfo) & 10.3 & 12.0 & \textbf{9.8} & 10.4 & -2.0 & \textbf{8.8} \\
 Methods & HUML-ensemble & \textbf{10.2} & \textbf{12.8} & 9.9 & \textbf{10.2} & \textbf{0.0} & 8.9 \\ \hline 
Operational  & FSSE & \textbf{9.7} & \textbf{17.1} & \textbf{9.5} & \textbf{8.5} & \textbf{16.7} & \textbf{7.8} \\
 Forecasts & OFCL & 10.0 & 14.5 & 10.1 & \textbf{8.5} & \textbf{16.7} & 8.1 \\ \hline 
Consensus & OP-average consensus & 9.6 & 17.9 & 9.7 & 8.5 & 16.7 & 7.9 \\
 Models & HUML/OP-average consensus & \textbf{9.2} & \textbf{21.4} & \textbf{9.0} & \textbf{8.3} & \textbf{18.6} & \textbf{7.6}\\
\hline
\end{tabular}
\label{table:cons_int}
\end{table*}

\paragraph{A multimodal approach leads to more accurate forecasts than using single data sources.\\}

As shown in Tables \ref{table:standalone_trk} and \ref{table:standalone_int}, for both track and intensity forecasts, multimodal models achieve higher accuracy and lower standard deviation than the model using only statistical data. 

The deep-learning feature extraction methods outperform the tensor-decomposition-based approach. 
This is not surprising as our encoder-decoders trained with a supervised learning objective, which means extracted features are tailored for the particular downstream prediction task. Tensor decomposition is, however, advantageously label-agnostic but did not extract features with enough predictive information to improve the performance.




\section{Limitations and Extensions} \label{sec:discussions}

\subsection{The Use of Reanalysis Maps} \label{sec:real-world}

A significant limitation of reanalysis maps is the computation time for construction, as they are assimilated based on observational data. Thus, although our models can compute forecasts in seconds, the dependence on reanalysis maps is a bottleneck in real-time forecasting. Therefore, a natural extension for effective deployment is to train our models using real-time observational data or field forecasts from powerful dynamical models such as HWRF. 
Since dynamical models are constantly updated with improved physics, higher resolution, and fixed bugs, reforecast products (e.g., \cite{hamill}) should be well-suited for training our encoder-decoders.
Nevertheless, we hope our framework could provide guidance and reference to build operational machine learning models in the future. 
\subsection{Incorporate Additional Data}

Under the scope of this work, we used nine reanalysis maps per time step, corresponding to the geopotential height ($z$), the zonal ($u$) and meridional ($v$) component of the wind fields from three pressure levels. 
One natural extension is to include additional features, such as the sea-surface temperature, the temperature, and the relative humidity, and include information from more vertical levels to potentially improve model performance. 

In addition, one could include more data sources, such as satellite and radar data.
Notably, we highlight the flexibility of our framework that can easily incorporate new data: we can adopt different feature extraction architectures and then append or substitute extracted features in the XGBoost forecasting model accordingly.

\subsection{Longer-Term Forecasts}

We conducted our experiments for 24-hour lead time predictions to demonstrate the potential of ML techniques in hurricane forecasting tasks. However, experiments on longer-term forecasts are needed before deploying such approaches. For example, the official NHC forecast provides guidance for up to 5 days. Nevertheless, our framework can be extended to longer lead-time forecasts. In particular, we recommend extending the input window size (from current 24-hour) as our models can process arbitrary long input sequences. 

\section{Conclusion}

This study demonstrates a novel multimodal machine learning framework for tropical cyclone intensity and track forecasting utilizing historical storm data and meteorological reanalysis data. We present a three-step pipeline to combine multiple machine learning approaches, consisting of (1) deep feature extraction, (2) concatenation of all processed features, (3) prediction.  We demonstrate that a successful combination of deep learning techniques and gradient-boosted trees can achieve strong predictions for both track and intensity forecasts, producing comparable results to current operational forecast models, especially in the intensity task. We acknowledge that the unavailability of real-time reanalysis data poses a challenge for operational use, and suggest future work to extend our framework with other operational data sources. 

We demonstrate that multimodal encoder-decoder architectures can successfully serve as a spatial-temporal feature extractor for downstream prediction tasks. In particular, this is also the first successful application of a Transformer-decoder architecture in tropical cyclone forecasting.

Furthermore, we show that consensus models that include our machine learning model could benefit the NHC's official forecast for both intensity and track, thus demonstrating the potential value of developing machine learning approaches as a new branch methodology for tropical cyclone forecasting.

Moreover, once trained, our models run in seconds, showing practical interest for real-time forecast, the bottleneck lying only in the data acquisition. We propose extensions and guidance for effective real-world deployment.

In conclusion, our work demonstrates that machine learning can provide valuable additions to the field of tropical cyclone forecasting. We hope this work opens the door for further use of machine learning in meteorological forecasting.

\acknowledgments

We thank the review team of the Weather and Forecasting journal for insightful comments that improved the paper substantially. We thank Louis Maestrati, Sophie Giffard-Roisin, Charles Guille-Escuret, Baptiste Goujaud, David Yu-Tung Hui, Ding Wang, Tianxing He for useful discussions. We thank Nicol\`o Forcellini, Miya Wang for proof-reading. The work was partially supported from a grant to MIT by the OCP Group. The authors acknowledge the MIT SuperCloud and Lincoln Laboratory Supercomputing Center for providing high-performance computing resources that have contributed to the research results reported within this paper.


%
%
\datastatement

All the data we used is open-source and can directly be accessed from the Internet with IBTrACS for TC features, Tropycal for operational forecasts, ERA-5 for vision data. Our code is available at \url{https://github.com/leobix/hurricast}.

%





\newpage 
\appendix \label{app:tensor}

\appendixtitle{Technical Components}

\section{Encoder-Decoder Architectures}

\subsection{Overall Architecture and Mechanisms}

\paragraph*{The CNN-encoder}
At each time step, we feed the nine reanalysis maps into the CNN-encoder, which produces one-dimensional embeddings. The CNN-encoder consists of three convolutional layers, with ReLU activation and MaxPool layers in between, followed by two fully connected layers. 

Next, we concatenate the reanalysis maps embeddings with processed statistical data corresponding to the same time step. At this point, data is still sequentially structured as 8 time steps to be passed on to the decoder. 

\paragraph*{The GRU-Decoder} 
Our GRU-decoder consists of two unidirectional layers. The data sequence embedded by the encoder is fed sequentially in chronological order into the GRU-decoder. For each time step, the GRU-decoder outputs a hidden state representing a ``memory'' of the previous time steps. Finally, a track or intensity prediction is made based upon these hidden states concatenated all together and given as input to fully-connected layers (see Figure \ref{fig:cnn_gru}).

\paragraph*{The Transformer-Decoder}

Conversely to the GRU-decoder, we feed the sequence as a whole into the Transformer-decoder. The time-sequential aspect is lost since attention mechanisms allow each hidden representation to attend holistically to the other hidden representations. Therefore, we add a \textit{positional encoding} token at each timestep-input, following standard practices \citep{transformers}. This token represents the relative position of a time-step within the sequence and re-introduces some information about the inherent sequential aspect of the data and experimentally improves performance.

Then, we use two Transformer layers that transform the 8 time steps (of size 142) into an 8-timestep sequence with similar dimensions. To obtain a unique representation of the sequence, we average the output sequence feature-wise into a one-dimensional vector, following standard practices. Finally, a track or intensity prediction is made based upon this averaged vector input into one fully-connected layer (see Figure \ref{fig:cnn_transfo}).

\paragraph*{Loss function}

The network is trained using an objective function $\mathcal{L}$ based on a mean-squared-error loss on the variable of interest (maximum sustained wind speed or TC displacement) added to an $L2$ regularization term on the weights of the network:
\[\mathcal{L}:=\frac{1}{N} \sum_{i=1}^{N}\left(y_{i}^{\text{true}}-y_{i}^{\text{pred}}\right)^2 + \lambda \sum_{l} \sum_{k,j} W_{k, j}^{[l] 2}, \]
where $N$ is the number of predictions, $y_{i}^{\text{pred}}$ the predicted forecast intensity or latitude-longitude displacements with a lead time of 24 h, $y_{i}^{\text{true}}$
the ground-truth values, $\lambda$ a regularization parameter chosen by validation, $W^{[l]}$ the weights of the $l$-th layer of the network. 
We minimize this loss function using the Adam optimizer \citep{adam}.

\subsection{Technical Details on the CNN-Encoder GRU-Decoder Network}

We provide more formal and precise explanations of our encoder-decoder architectures.

\paragraph{CNN-encoder GRU-decoder architecture details}

Let $t$ the instant when we want to make a 24-hour lead time prediction. Let $\textbf{x}^{\text{viz}}_t \in \mathbb{R}^{8 \times 9 \times 25 \times 25}$ be the corresponding spatial-temporal input of the CNN, where 8 is the number of past time steps in the sequence, 9 is the number of pressure levels times the number of features maps, $25^{\circ} \times 25^{\circ}$ is the pixel size of each reanalysis map. Let $\textbf{x}^{\text{stat}}_t \in \mathbb{R}^{8 \times 31}$ be the corresponding statistical data, where 8 is the number of time steps in the sequence, and 31 the number of features available at each time step.

First, $\textbf{x}^{\text{viz}}_t$ is embedded by the CNN into $\textbf{x}^{\text{emb}}_t \in \mathbb{R}^{8 \times 128}$ where 8 is the number of time steps in the sequence, 128 is the dimension of the embedding space. Figure A1 provides an illustration of this embedding process by the CNN-encoder.

Let $i \in \{0,\dots,7\}$ be the corresponding index of the time step in the sequence $t$. At each time step $t_i$ of the sequence, the CNN embedding $\textbf{x}^{\text{emb}}_{t_i}$ is concatenated with the statistical data $\textbf{x}^{\text{stat}}_{t_i}$ and processed as \[\textbf{h}_{t_i} := \text{GRU}(\textbf{h}_{t_{i-1}},  [\textbf{x}^{\text{emb}}_{t_i}, \textbf{x}^{\text{stat}}_{t_i}]),\] with $\textbf{h}_{t_0} =$ \textbf{0}, $\textbf{h}_{t_i} \in \mathbb{R}^{128}, \forall i$. 
$\quad [\cdot , \cdot]$ means concatenation of the two vectors along the column axis, to keep a one-dimensional vector. 

Finally, we concatenate $\textbf{h}_{t_{0}}, \textbf{h}_{t_{1}}, \dots, \textbf{h}_{t_{7}}$ to obtain a one-dimensional vector $\textbf{x}^{\text{hidden}}_t$ of size $8 \cdot 128 = 1024$ and pass this vector into a series of 3 fully connected linear layers, of input-output size: (1024, 512); (512, 128); (128,$c$), where $c = 2$ for track forecast task and and $c = 1$ for intensity task. The final layer makes the prediction.

To extract the spatial-temporal embedded features, we use the output of the second fully connected layer, of dimension 128. Therefore, this technique allows to reduce $8 \cdot 9 \cdot 25 \cdot 25 = 45,000$ features into 128 predictive features that can be input into our XGBoost models.

\begin{figure}[h]
    \includegraphics[width=1\linewidth]{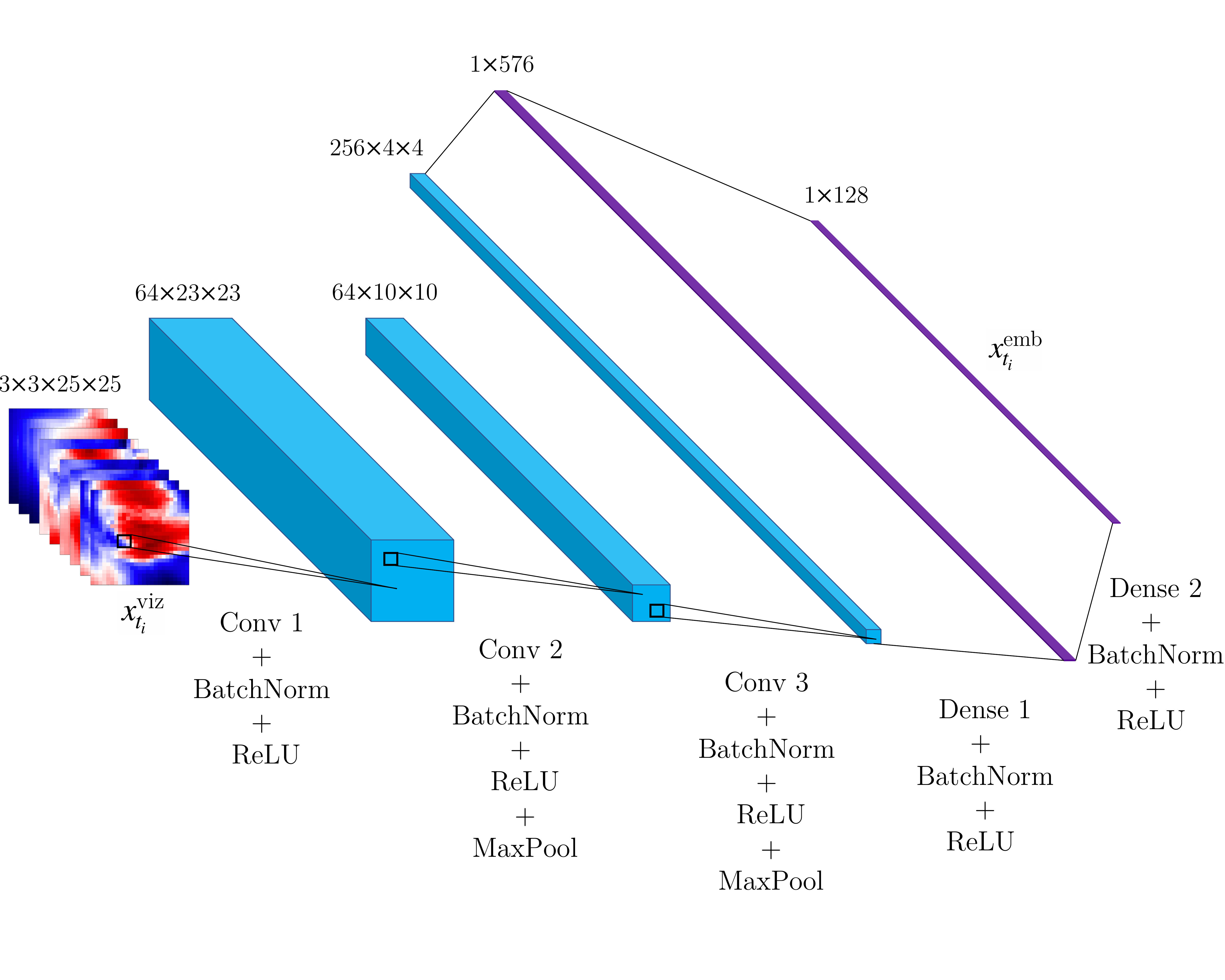}
    \appendcaption{A1}{Representation of our CNN-encoder. We use 3 convolutional layers, with batch normalization, ReLU and MaxPool in between. We use fully connected (dense) layers to obtain in the end a one-dimensional vector $x_{t_i}^{\text{emb}}$. }
\end{figure}

For each convolutional layer of the CNN, we use the following parameters: kernel size = 3, stride = 1, padding = 0.
For each MaxPool layer, we use the following parameters: kernel size = 2, stride = 2, padding = 0.

The CNN-encoder architecture is inspired from \cite{sophie}. The combination with the GRU-decoder or Transformer-decoder and the feature extraction is a contribution of our work.

\subsection{Technical Details on the Transformer-Decoder Architecture}

As with the CNN-encoder GRU-decoder network, the spatial-temporal inputs are processed and concatenated with the statistical data to obtain a sequence of input $[\textbf{x}_{t_i}^{emb}, \textbf{x}_{t_i}^{stat}], \forall i \in \{0,...,7\}$. As suggested by \cite{transformers}, we add to each $[\textbf{x}_{t_i}^{emb}, \textbf{x}_{t_i}^{stat}]$ input a positional encoding $\textbf{P}_i$ token in order to provide some information about the relative position $i$ within the sequence. We eventually obtain $\textbf{x} = [\textbf{x}_{t_i}^{emb}, \textbf{x}_{t_i}^{stat}] + \textbf{P}_i$ which is being processed by the Transformer's layers. 
In this work, we use $P_{i,2j} = \sin(i/10000^{2j/d})$and  $P_{i, 2j+1} = \cos(i/10000^{2j/d})$, where $i$ is the position in the sequence, $j$ the dimension and $d$ the dimension of the model, in our case 142.
A layer is composed of a multi-head attention transformation followed by a fully-connected layer, similar to the Transformer's encoder presented in \cite{transformers}.

We used self-attention layers (i.e., $Q=K=V$), specifically 2 layers with 2 heads, the model's dimension $d_k$ being fixed to 142 and the feedforward dimension set to 128. 

We then averaged the outputs of our Transformer $\textbf{h}_{t_0}, \dots, \textbf{h}_{t_7}$ feature-wise to obtain the final representation of the sequence.

\section{Tucker Decomposition for Tensors}

The multilinear singular value decomposition (SVD) expresses a tensor $\mathcal{A}$ as a small core tensor $\mathcal{S}$  multiplied by a set of unitary matrices. The size of the core tensor, denoted by $[k_1, \dots k_N]$, defines the rank of the tensor. 
 
Formally, the multilinear decomposition can be expressed as:
\begin{align*}
  \mathcal{A} &= \mathcal{S} \times_1 U ^{(1)} \times_2 \dots \times_N U^{(N)}\\ 
\text{where  }
     \mathcal{A} &\in  \mathbb{R}^{I_1 \times I_2 \times \dots \times I_N} \\
     \mathcal{S} &\in  \mathbb{R}^{k_1 \times k_2 \times \dots \times k_N} \\
      U ^{(i)} &\in \mathbb{R}^{I_i \times k_i} 
\end{align*} 
where each $U^{(i)}$ is a unitary matrix, i.e., its conjugate transpose is its inverse $U^{(i)*}U^{(i)} = U^{(i)}U^{(i)*} = I$, and the mode-n product, denoted by $\mathcal{A} \times_n U$, denotes the multiplication operation of a tensor $\mathcal{A} \in  \mathbb{R}^{I_1 \times I_2 \times \dots \times I_N} $ by a matrix $ U \in  \mathbb{R}^{I_n \times J_n} $. Figure A2 exhibits a geometric representation of the Tucker decomposition applied to a three-dimensional tensor $\mathcal{A}$, which is decomposed as a smaller core tensor $\mathcal{S}$ and projection maps $U^{i}_{i=1,2,3}$. 

\begin{figure}[h]
    \includegraphics[width=0.95\linewidth]{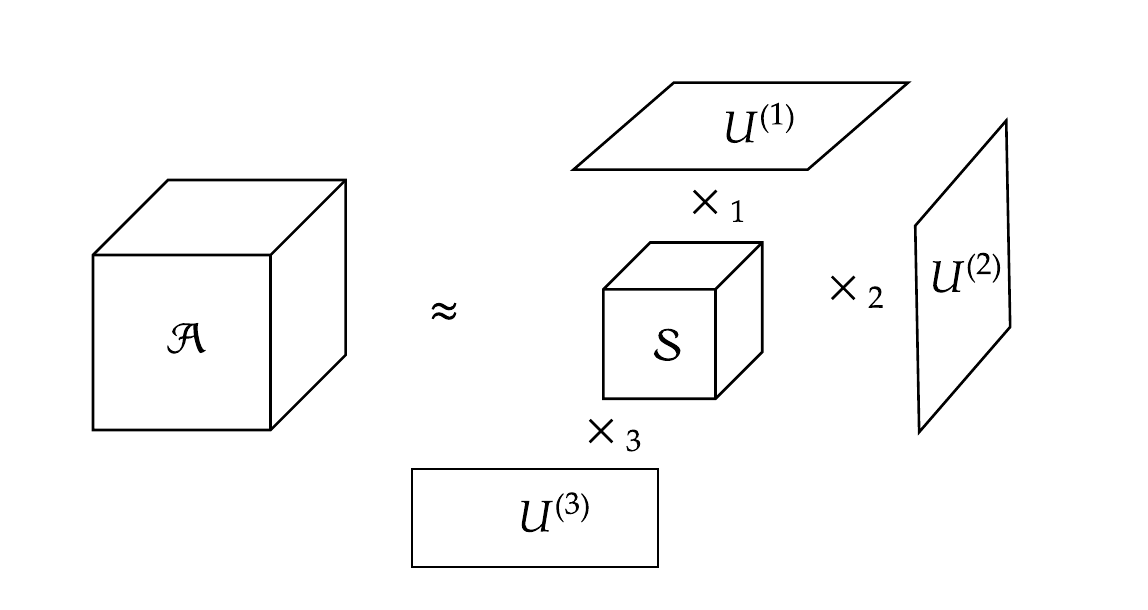}
    \appendcaption{A2}{Illustration of the tensor decomposition of a 3 dimensional tensor. Tensor $\mathcal{A}$ is the original tensor, which is approximated through Tucker decomposition using a core tensor tensor $\mathcal{S}$ and three linear projection maps along each axis $U^{(1)}, U^{(2)}, U^{(3)}$.}
\end{figure}

Analogous to truncated SVD, we can reduce the dimensionality of tensor $\mathcal{A}$ by artificially truncating the core tensor $\mathcal{S}$ and corresponding $U^{(i)}$. For instance, given a 4-dimensional tensor of TC maps, we can decide to reduce the tensor to any desired rank by keeping only the desired size of core tensor $\mathcal{S}$. For instance, to reduce TC tensor data into rank $3\times 5\times 3\times 3$, we first perform multilinear SVD, such that S reflects descending order of the singular values, and then truncate $\mathcal{S}$ by keeping only the first $3\times 5\times 3\times 3$ entries, denoted by $\mathcal{S}'$, and the first 3 columns of each of $U^{(i)}$, denoted by $U'^{(i)}$. 

 Finally, we flatten the truncated core tensor $\mathcal{S}'$ into a vector, which is treated as the extracted vision features in order to train the XGBoost model.

\section{Experiment Details}

\subsection{Testing Methodology}\label{sec:testing}

We employed the validation set to perform hyperparameter tuning. Then, we retrained the models on the training and validation set combined using the best combination of hyperparameters. We then evaluate our models' performance on the test set.

We report the performance obtained on the NA and EP test set with each method for 24-hour lead time for both intensity and track forecasts. As a remark, in reality, there is often a time lag when operational models become available. Such lag is shorter for statistical models but longer for dynamical models (up to several hours) because of expensive computational time. Due to the lag time variability, we do not consider such lag in our comparisons with operational models. In other words, we neglect the time lag for all models and compare model results assuming all forecasts compute instantaneously. We hope to provide an overall sense of the predictive power of our methodology, although we acknowledge that using reanalysis maps data is not possible in real-time. We discussed this bottleneck in section  \ref{sec:discussions}.

\subsection{The Specific Protocol for HUML-ensemble}

For the HUML-ensemble model, we used the HUML models 1-4 trained on the training set only (i.e., data until 2011). We then used their forecasts on the unseen validation set  (2012 to 2015) and their forecasts on the unseen test set (2016 to 2019) as the training and testing data for the ensemble. The goal is to understand how each model behaves with respect to the others on unseen data. We cross-validated the ElasticNet parameters on the 2012-2015 HUML forecasts and we finally tested on the same cases as before using the best hyperparameter combination found.

\subsection{Hyperparameter Tuning}\label{hyperparam}

We distinguish five categories of hyperparameters to tune: (1) the data-related features, (2) the neural-network related features, (3) the tensor decomposition-related features, (4) the tree-based method related features, (5) the consensus models-related features. 

\subsubsection{Data-related features}

The data-related features include the area covered by the reanalysis maps (grid size) and the number of historical time steps of data to use for each forecast. We tune these features by comparing the 24-hour lead time forecast performance of the encoder-decoders for each different hyperparameter configuration. 

We found that using eight past time steps (i.e., up to 21 hours in the past) and a grid size of $25 \times 25$ degrees for the reanalysis maps was the best combination. We also found that standardizing the vision and statistical data --- i.e., rescaling each feature to mean 0 and standard deviation 1 --- yielded better results than normalizing --- i.e., rescaling each feature to the $[0,1]$ range.

\subsubsection{Neural network-related features} 

The neural network-related features include the optimizer, the architecture itself, the batch size during training, and the loss function's regularizer. 

The best results were obtained using a batch size of 64, a $\lambda$ regularization term of 0.01, and the encoder-decoder architectures described previously.
Regarding the optimizer, we use Adam \citep{adam} with a learning rate of $10^{-3}$ for the intensity forecast and $4 \cdot 10^{-4}$ for the track forecast.

\subsubsection{Tensor decomposition features}
The tensor decomposition algorithm includes the choice of the core tensor size, i.e., the compressed size of the original tensor. Recall that the original tensor size is $8\times 9 \times 25\times 25$. Based on empirical testing, we found using a small tensor size of $ 3\times 5 \times 3\times 3$ yielded the best performance when compressed reanalysis maps are included as features in XGBoost models.  

\subsubsection{Tree-based method features} 

Based on empirical testing, we found XGBoost models consistently outperforming Decision Trees and Random Forests or other ML methods such as Support Vector Machines, Regularized Linear Regression and Multi-Layer Perceptrons. XGBoost trains also fast which is a considerable advantage for heavy hyperparameter search. Therefore, we selected XGBoost as the core model for prediction. 

Then, there is variability in the best combinations of hyperparameters depending on each task (track or intensity), basin (NA or EP) or data sources to use (statistical, various reanalysis maps embeddings). However, these particular features were typically important and were the best in the following ranges: maximum depth of the trees (between 6 and 9), number of estimators (between 100 and 300), learning rate (between 0.03 and 0.15), subsample (between 0.6 and 0.9), column sampling by tree (between 0.7 and 1), minimum child by tree (between 1 and 5).

\subsubsection{Consensus-models-related features} 

We tested different kinds of consensus models on the HUML forecasts, including ElasticNet \citep{elastic}, tree-based models, and multi-layer perceptrons (MLPs) as meta-learners. MLPs had similar performance with ElasticNet, but since they are less interpretable and stable, ElasticNet is the strongest ensembler candidate and our final choice for HUML-ensemble. 
We tune the L1/L2 ratio between 0 and 1 and the regularization penalty between $10^{-4}$ and 10.

\subsection{Metrics}

\subsubsection{Haversine Formula}

Formally, the Haversine distance between one pair of predicted point and actual point, denoted by $d$, is calculated by:
\begin{align*}
d&=2 R \arcsin \left(\sqrt{\alpha}\right),  \text{ where   }\\
\alpha&= \sin ^{2}\left(\frac{\hat{\phi}-\phi}{2}\right)+\cos \left(\hat{\phi}\right) \cos \left(\phi\right) \sin ^{2}\left(\frac{\hat{\lambda}-\lambda}{2}\right)
\end{align*}
where $(\phi,\lambda)$ are the actual latitude and longitude of one data point,  $(\hat{\phi},\hat{\lambda})$  are the predicted latitude and longitude, and $R$ is Earth’s radius, approximated to be the mean radius at 6,371 km.

\subsubsection{Skill}

Skill represents a normalization of the forecast error against a standard or baseline. We computed the skill $s_f$ of a forecast $f$ following \citep{nhc2020doc}:
\[s_f (\%)=100 \cdot \frac{e_b - e_f}{e_b}\]
where $e_b$ is the error of the baseline model and $e_f$ is the error of the forecast being evaluated. Skill is positive when the forecast error is smaller than the error from the baseline.

\bibliographystyle{ametsoc2014}
\bibliography{references}

\begin{thebibliography}{48}
\providecommand{\natexlab}[1]{#1}
\providecommand{\url}[1]{\texttt{#1}}
\renewcommand{\UrlFont}{\rmfamily}
\providecommand{\urlprefix}{URL }
\expandafter\ifx\csname urlstyle\endcsname\relax
  \providecommand{\doi}[1]{doi:\discretionary{}{}{}#1}\else
  \providecommand{\doi}{doi:\discretionary{}{}{}\begingroup
  \urlstyle{rm}\Url}\fi
\providecommand{\eprint}[2][]{\url{#2}}

\bibitem[{Aberson(1998)}]{cliper5}
Aberson, S.~D., 1998: Five-day tropical cyclone track forecasts in the north
  atlantic basin. \textit{Weather and Forecasting}, \textbf{13~(4)}, 1005 --
  1015.

\bibitem[{Alemany et~al.(2019)Alemany, Beltran, Perez,, and
  Ganzfried}]{rnn2019}
Alemany, S., J.~Beltran, A.~Perez, and S.~Ganzfried, 2019: Predicting hurricane
  trajectories using a recurrent neural network. \textit{Proceedings of the
  AAAI Conference on Artificial Intelligence}, Vol.~33, 468--475.

\bibitem[{Bahdanau et~al.(2015)Bahdanau, Cho,, and Bengio}]{bahdanau2014neural}
Bahdanau, D., K.~Cho, and Y.~Bengio, 2015: Neural machine translation by
  jointly learning to align and translate. \textit{CoRR},
  \textbf{abs/1409.0473}.

\bibitem[{Bian et~al.(2021)Bian, Nie,, and Qiu}]{bian}
Bian, G.-F., G.-Z. Nie, and X.~Qiu, 2021: How well is outer tropical cyclone
  size represented in the era5 reanalysis dataset? \textit{Atmospheric
  Research}, \textbf{249}, 105\,339.

\bibitem[{Biswas et~al.(2018)}]{hwrf2018doc}
Biswas, M.~K., and Coauthors, 2018: Hurricane weather research and forecasting
  (hwrf) model: 2018 scientific documentation. \textit{Developmental Testbed
  Center}.

\bibitem[{Burg and Lillo(2020)Burg, and Lillo}]{tropycal}
Burg, T., and S.~P. Lillo, 2020: Tropycal: A python package for analyzing
  tropical cyclones and more. \textit{34th Conference on Hurricanes and
  Tropical Meteorology}, AMS.

\bibitem[{Cangialosi(2020)}]{nhc2020doc}
Cangialosi, J.~P., 2020: National hurricane center forecast verification
  report. \textit{National Hurricane Center},
  \urlprefix\url{https://www.nhc.noaa.gov/verification/pdfs/Verification_2020.pdf}.

\bibitem[{Cangialosi et~al.(2020)Cangialosi, Blake, DeMaria, Penny, Latto,
  Rappaport,, and Tallapragada}]{nhc}
Cangialosi, J.~P., E.~Blake, M.~DeMaria, A.~Penny, A.~Latto, E.~Rappaport, and
  V.~Tallapragada, 2020: {Recent Progress in Tropical Cyclone Intensity
  Forecasting at the National Hurricane Center}. \textit{Weather and
  Forecasting}, 1--30.

\bibitem[{Chen et~al.(2019)Chen, Wang, Zhang, Zhu, Li,, and Yang}]{springer}
Chen, R., X.~Wang, W.~Zhang, X.~Zhu, A.~Li, and C.~Yang, 2019: A hybrid
  cnn-lstm model for typhoon formation forecasting. \textit{GeoInformatica}.

\bibitem[{Chen and Guestrin(2016)Chen, and Guestrin}]{xgboost}
Chen, T., and C.~Guestrin, 2016: Xgboost: A scalable tree boosting system.
  \textit{Proceedings of the 22nd ACM SIGKDD International Conference on
  Knowledge Discovery and Data Mining}, ACM, 785--794, KDD '16.

\bibitem[{Chung et~al.(2014)Chung, G{\"{u}}l{\c{c}}ehre, Cho,, and
  Bengio}]{gru}
Chung, J., {\c{C}}.~G{\"{u}}l{\c{c}}ehre, K.~Cho, and Y.~Bengio, 2014:
  Empirical evaluation of gated recurrent neural networks on sequence modeling.
  \textit{CoRR}, \textbf{abs/1412.3555}, \eprint{1412.3555}.

\bibitem[{De~Lathauwer et~al.(2000)De~Lathauwer, De~Moor,, and
  Vandewalle}]{tensor_multisvd}
De~Lathauwer, L., B.~De~Moor, and J.~Vandewalle, 2000: A multilinear singular
  value decomposition. \textit{SIAM journal on Matrix Analysis and
  Applications}, \textbf{21~(4)}, 1253--1278.

\bibitem[{DeMaria and Kaplan(1994)DeMaria, and Kaplan}]{SHIPS1994model}
DeMaria, M., and J.~Kaplan, 1994: A statistical hurricane intensity prediction
  scheme (ships) for the atlantic basin. \textit{Weather and Forecasting},
  \textbf{9~(2)}, 209--220.

\bibitem[{DeMaria et~al.(2005)DeMaria, Mainelli, Shay, Knaff,, and
  Kaplan}]{shipsdemaria}
DeMaria, M., M.~Mainelli, L.~K. Shay, J.~A. Knaff, and J.~Kaplan, 2005: Further
  improvements to the statistical hurricane intensity prediction scheme
  (ships). \textit{Weather and Forecasting}, \textbf{20~(4)}, 531 -- 543.

\bibitem[{ECWMF(2019)}]{ecmwf2019doc}
ECWMF, 2019: \textit{PART III: Dynamics and Numerical Procedures}. No.~3, IFS
  Documentation, ECMWF, \urlprefix\url{https://www.ecmwf.int/node/19307}.

\bibitem[{Gao et~al.(2018)Gao, Zhao, Pan, Li, Zhou, Xu, Zhong,, and Shi}]{gao}
Gao, S., P.~Zhao, B.~Pan, Y.~Li, M.~Zhou, J.~Xu, S.~Zhong, and Z.~Shi, 2018: A
  nowcasting model for the prediction of typhoon tracks based on a long short
  term memory neural network. \textit{Acta Oceanologica Sinica}, \textbf{37},
  8--12.

\bibitem[{Giffard-Roisin et~al.(2020)Giffard-Roisin, Yang, Charpiat,
  Kumler~Bonfanti, Kégl,, and Monteleoni}]{sophie}
Giffard-Roisin, S., M.~Yang, G.~Charpiat, C.~Kumler~Bonfanti, B.~Kégl, and
  C.~Monteleoni, 2020: Tropical cyclone track forecasting using fused deep
  learning from aligned reanalysis data. \textit{Frontiers in Big Data},
  \textbf{3}, 1.

\bibitem[{Goodfellow et~al.(2016)Goodfellow, Bengio,, and
  Courville}]{deeplearningbook}
Goodfellow, I., Y.~Bengio, and A.~Courville, 2016: \textit{Deep Learning}. The
  MIT Press.

\bibitem[{Grinsted et~al.(2019)Grinsted, Ditlevsen,, and
  Christensen}]{grinsted}
Grinsted, A., P.~Ditlevsen, and J.~H. Christensen, 2019: Normalized us
  hurricane damage estimates using area of total destruction, 1900-2018.
  \textit{Proceedings of the National Academy of Sciences}, \textbf{116~(48)},
  23\,942--23\,946.

\bibitem[{Hamill et~al.(2013)Hamill, Bates, Whitaker, Murray, Fiorino,
  Galarneau, Zhu,, and Lapenta}]{hamill}
Hamill, T., G.~Bates, J.~Whitaker, D.~Murray, M.~Fiorino, T.~Galarneau, Y.~Zhu,
  and W.~Lapenta, 2013: Noaa's second-generation global medium-range ensemble
  reforecast dataset. \textit{Bulletin of the American Meteorological Society},
  \textbf{94}, 1553--1565.

\bibitem[{Harper et~al.(2010)Harper, Kepert,, and Ginger}]{guidelines}
Harper, B., J.~Kepert, and J.~Ginger, 2010: Guidelines for converting between
  various wind averaging periods in tropical cyclone conditions.
  \textit{Geneva, Switzerland: WMO, 2010},
  \urlprefix\url{https://library.wmo.int/doc_num.php?explnum_id=290}.

\bibitem[{He et~al.(2016)He, Zhang, Ren,, and Sun}]{deepnet2016}
He, K., X.~Zhang, S.~Ren, and J.~Sun, 2016: Deep residual learning for image
  recognition. \textit{Proceedings of the IEEE conference on computer vision
  and pattern recognition}, 770--778.

\bibitem[{Hersbach et~al.(2020)}]{era5_good}
Hersbach, H., and Coauthors, 2020: The era5 global reanalysis.
  \textit{Quarterly Journal of the Royal Meteorological Society},
  \textbf{146~(730)}, 1999--2049.

\bibitem[{Hinton et~al.(2015)Hinton, Vinyals,, and Dean}]{distillation}
Hinton, G., O.~Vinyals, and J.~Dean, 2015: Distilling the knowledge in a neural
  network. \textit{NIPS Deep Learning and Representation Learning Workshop}.

\bibitem[{Hodges et~al.(2017)Hodges, Cobb,, and Vidale}]{hodges}
Hodges, K., A.~Cobb, and P.~L. Vidale, 2017: How well are tropical cyclones
  represented in reanalysis datasets? \textit{Journal of Climate},
  \textbf{30~(14)}, 5243 -- 5264.

\bibitem[{Kiela and Bottou(2014)Kiela, and Bottou}]{transfer1}
Kiela, D., and L.~Bottou, 2014: Learning image embeddings using convolutional
  neural networks for improved multi-modal semantics. \textit{Proceedings of
  the 2014 Conference on Empirical Methods in Natural Language Processing
  ({EMNLP})}, Association for Computational Linguistics, 36--45.

\bibitem[{Kingma and Ba(2014)Kingma, and Ba}]{adam}
Kingma, D., and J.~Ba, 2014: Adam: A method for stochastic optimization.
  \textit{International Conference on Learning Representations}.

\bibitem[{Knaff et~al.(2003)Knaff, DeMaria, Sampson,, and Gross}]{SHIFOR5}
Knaff, J., M.~DeMaria, C.~Sampson, and J.~Gross, 2003: Statistical 5-day
  tropical cyclone intensity forecasts derived from climatology and
  persistence. \textit{Weather and Forecasting}, \textbf{18}, 80 --92.

\bibitem[{Knapp et~al.(2010)Knapp, Kruk, Levinson, Diamond,, and
  Neumann}]{ibtracs}
Knapp, K.~R., M.~C. Kruk, D.~H. Levinson, H.~J. Diamond, and C.~J. Neumann,
  2010: The international best track archive for climate stewardship (ibtracs):
  Unifying tropical cyclone best track data. Bulletin of the American
  Meteorological Society, 91, 363-376.

\bibitem[{Krizhevsky et~al.(2012)Krizhevsky, Sutskever,, and
  Hinton}]{imagenet2012}
Krizhevsky, A., I.~Sutskever, and G.~E. Hinton, 2012: Imagenet classification
  with deep convolutional neural networks. \textit{Advances in neural
  information processing systems}, 1097--1105.

\bibitem[{LeCun et~al.(1989)LeCun, Boser, Denker, Henderson, Howard, Hubbard,,
  and Jackel}]{cnn1989lecun}
LeCun, Y., B.~Boser, J.~S. Denker, D.~Henderson, R.~E. Howard, W.~Hubbard, and
  L.~D. Jackel, 1989: Backpropagation applied to handwritten zip code
  recognition. \textit{Neural computation}, \textbf{1~(4)}, 541--551.

\bibitem[{Lian et~al.(2020)Lian, Dong, Zhang, Pan,, and Liu}]{lian2020cnntrack}
Lian, J., P.~Dong, Y.~Zhang, J.~Pan, and K.~Liu, 2020: A novel data-driven
  tropical cyclone track prediction model based on cnn and gru with
  multi-dimensional feature selection. \textit{IEEE Access}.

\bibitem[{Moradi~Kordmahalleh et~al.(2016)Moradi~Kordmahalleh,
  Gorji~Sefidmazgi,, and Homaifar}]{rnn2016}
Moradi~Kordmahalleh, M., M.~Gorji~Sefidmazgi, and A.~Homaifar, 2016: A sparse
  recurrent neural network for trajectory prediction of atlantic hurricanes.
  \textit{Proceedings of the Genetic and Evolutionary Computation Conference
  2016}, 957--964.

\bibitem[{Mudigonda et~al.(2017)}]{mudigonda}
Mudigonda, M., and Coauthors, 2017: Segmenting and tracking extreme climate
  events using neural networks.

\bibitem[{{National Hurricane Center}(2021)}]{atcf}
{National Hurricane Center}, 2021: Automated tropical cyclone forecasting
  system (atcf). National Hurricane Center (NHC),
  \urlprefix\url{https://ftp.nhc.noaa.gov/atcf/}, accessed: 2021-04-06.

\bibitem[{Paszke et~al.(2019)}]{pytorch}
Paszke, A., and Coauthors, 2019: Pytorch: An imperative style, high-performance
  deep learning library. \textit{Advances in Neural Information Processing
  Systems 32}, H.~Wallach, H.~Larochelle, A.~Beygelzimer, F.~d\textquotesingle
  Alch\'{e}-Buc, E.~Fox, and R.~Garnett, Eds., Curran Associates, Inc.,
  8024--8035.

\bibitem[{Rumelhart et~al.(1985)Rumelhart, Hinton,, and
  Williams}]{rumelhart1985learning}
Rumelhart, D.~E., G.~E. Hinton, and R.~J. Williams, 1985: Learning internal
  representations by error propagation. Tech. rep., California Univ San Diego
  La Jolla Inst for Cognitive Science.

\bibitem[{Sampson and Schrader(2000)Sampson, and Schrader}]{atcf2}
Sampson, C., and A.~J. Schrader, 2000: The automated tropical cyclone
  forecasting system (version 3.2). \textit{Bulletin of the American
  Meteorological Society}, \textbf{81}, 1231--1240.

\bibitem[{Sampson et~al.(2008)Sampson, Franklin, Knaff,, and
  DeMaria}]{consensusintensity}
Sampson, C.~R., J.~L. Franklin, J.~A. Knaff, and M.~DeMaria, 2008: Experiments
  with a simple tropical cyclone intensity consensus. \textit{Weather and
  Forecasting}, \textbf{23~(2)}, 304 -- 312.

\bibitem[{Schenkel and Hart(2012)Schenkel, and Hart}]{schenkel}
Schenkel, B.~A., and R.~E. Hart, 2012: An examination of tropical cyclone
  position, intensity, and intensity life cycle within atmospheric reanalysis
  datasets. \textit{Journal of Climate}, \textbf{25~(10)}, 3453 -- 3475.

\bibitem[{Shimada et~al.(2018)Shimada, Owada, Yamaguchi, Iriguchi, Sawada,
  Aonashi, DeMaria,, and Musgrave}]{demaria2005}
Shimada, U., H.~Owada, M.~Yamaguchi, T.~Iriguchi, M.~Sawada, K.~Aonashi,
  M.~DeMaria, and K.~D. Musgrave, 2018: {Further Improvements to the
  Statistical Hurricane Intensity Prediction Scheme Using Tropical Cyclone
  Rainfall and Structural Features}. \textit{Weather and Forecasting},
  \textbf{33~(6)}, 1587--1603.

\bibitem[{Simon et~al.(2018)Simon, Penny, DeMaria, Franklin, Pasch, Rappaport,,
  and Zelinsky}]{hcca}
Simon, A., A.~B. Penny, M.~DeMaria, J.~L. Franklin, R.~J. Pasch, E.~N.
  Rappaport, and D.~A. Zelinsky, 2018: A description of the real-time hfip
  corrected consensus approach (hcca) for tropical cyclone track and intensity
  guidance. \textit{Weather and Forecasting}, \textbf{33~(1)}, 37 -- 57.

\bibitem[{Su et~al.(2020)Su, Wu, Jiang, Pai, Liu, Zhai, Tavallali,, and
  DeMaria}]{nasa}
Su, H., L.~Wu, J.~H. Jiang, R.~Pai, A.~Liu, A.~J. Zhai, P.~Tavallali, and
  M.~DeMaria, 2020: Applying satellite observations of tropical cyclone
  internal structures to rapid intensification forecast with machine learning.
  \textit{Geophysical Research Letters}, \textbf{47~(17)}.

\bibitem[{Tan et~al.(2018)Tan, Sun, Kong, Zhang, Yang,, and Liu}]{transfer2}
Tan, C., F.~Sun, T.~Kong, W.~Zhang, C.~Yang, and C.~Liu, 2018: A survey on deep
  transfer learning. \textit{CoRR}, \textbf{abs/1808.01974},
  \eprint{1808.01974}.

\bibitem[{Van~Rossum and Drake~Jr(1995)Van~Rossum, and Drake~Jr}]{python}
Van~Rossum, G., and F.~L. Drake~Jr, 1995: \textit{Python tutorial}. Centrum
  voor Wiskunde en Informatica Amsterdam, The Netherlands.

\bibitem[{Vaswani et~al.(2017)Vaswani, Shazeer, Parmar, Uszkoreit, Jones,
  Gomez, Kaiser,, and Polosukhin}]{transformers}
Vaswani, A., N.~Shazeer, N.~Parmar, J.~Uszkoreit, L.~Jones, A.~N. Gomez, L.~u.
  Kaiser, and I.~Polosukhin, 2017: Attention is all you need. \textit{Advances
  in Neural Information Processing Systems 30}, 5998--6008.

\bibitem[{Yosinski et~al.(2014)Yosinski, Clune, Bengio,, and
  Lipson}]{transfer0}
Yosinski, J., J.~Clune, Y.~Bengio, and H.~Lipson, 2014: How transferable are
  features in deep neural networks? \textit{CoRR}, \textbf{abs/1411.1792}.

\bibitem[{Zou and Hastie(2005)Zou, and Hastie}]{elastic}
Zou, H., and T.~Hastie, 2005: Regularization and variable selection via the
  elastic net. \textit{Journal of the Royal Statistical Society, Series B},
  \textbf{67}, 301--320.

\end{thebibliography}

%

%




%
%
%

\end{document}